\documentclass{article}

\usepackage{PRIMEarxiv}

\usepackage[utf8]{inputenc} 
\usepackage[T1]{fontenc}    
\usepackage{hyperref}       
\usepackage{amsmath}
\usepackage[ruled,linesnumbered]{algorithm2e}
\usepackage{multirow}
\usepackage{cleveref}
\usepackage{url}            
\usepackage{booktabs}       
\usepackage{amsfonts}       
\usepackage{nicefrac}       
\usepackage{microtype}      
\usepackage{lipsum}
\usepackage{fancyhdr}       
\usepackage{subcaption}
\usepackage{graphicx}       
\graphicspath{{media/}}     

\begin{document}

\title{Refining Segmentation On-the-Fly: An Interactive Framework for Point Cloud Semantic Segmentation} 

\author{
  Peng Zhang, Ting Wu, Jinsheng Sun, Weiqing Li, Zhiyong Su \\
  Nanjing University of Science and Technology \\
  \texttt{\{zhangpeng,wuting00\}@njust.edu.cn,jssun67@163.com,\{li\_weiqing,su\}@njust.edu.cn} \\
}
\maketitle

\begin{abstract}
Existing interactive point cloud segmentation approaches primarily focus on the object segmentation, which aim to determine which points belong to the object of interest guided by user interactions.
This paper concentrates on an unexplored yet meaningful task, i.e., interactive point cloud semantic segmentation, which assigns high-quality semantic labels to all points in a scene with user corrective clicks.
Concretely, we presents the first interactive framework for point cloud semantic segmentation, named InterPCSeg, which seamlessly integrates with off-the-shelf semantic segmentation networks without offline re-training, enabling it to run in an on-the-fly manner.
To achieve online refinement, we treat user interactions as sparse training examples during the test-time.
To address the instability caused by the sparse supervision, we design a stabilization energy to regulate the test-time training process.
For objective and reproducible evaluation, we develop an interaction simulation scheme tailored for the interactive point cloud semantic segmentation task.
We evaluate our framework on the S3DIS and ScanNet datasets with off-the-shelf segmentation networks, incorporating interactions from both the proposed interaction simulator and real users.
Quantitative and qualitative experimental results demonstrate the efficacy of our framework in refining the semantic segmentation results with user interactions.
The source code will be publicly available.
\end{abstract}

\section{Introduction}
\label{sec:intro}
\begin{figure}
  \centering
  \begin{subfigure}[t]{0.3\linewidth}
    \includegraphics[width=1\linewidth]{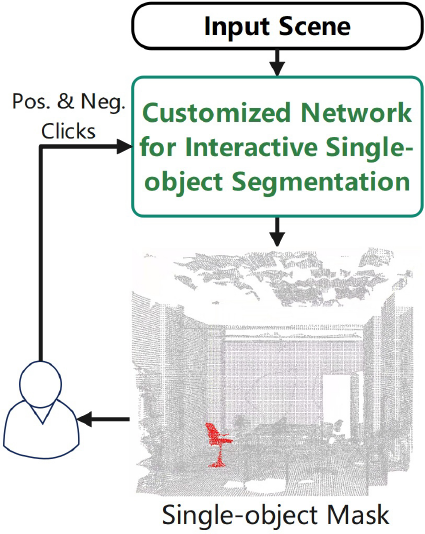}
    \caption{Customized approach for interactive point cloud single-object segmentation}
    \label{fig:intro_ISOS}
  \end{subfigure}
  \begin{subfigure}[t]{0.3\linewidth}
    \includegraphics[width=1\linewidth]{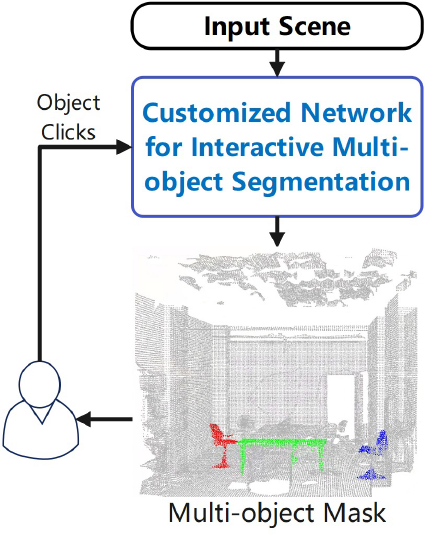}
    \caption{Customized approach for interactive point cloud multi-object segmentation}
    \label{fig:intro_IMOS}
  \end{subfigure}
  \begin{subfigure}[t]{0.302\linewidth}
    \includegraphics[width=1\linewidth]{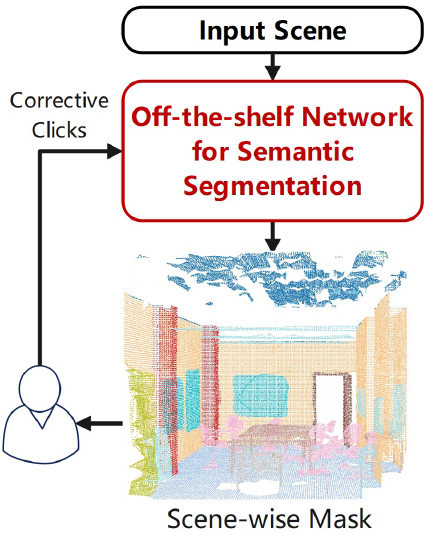}
    \caption{Our on-the-fly approach for interactive point cloud semantic segmentation}
    \label{fig:intro_ISS}
  \end{subfigure}
  \caption{Existing customized approaches depend on customized networks for object segmentation, which are decoupled with off-the-shelf networks and require offline training with tailored supervision.
  Our on-the-fly approach builds upon off-the-shelf semantic segmentation networks and only works at the test-time.}
  \label{fig:intro}
\end{figure}

Deep learning techniques have shown great potential in scene-level point cloud semantic segmentation \cite{qi-2017-cvpr,wang-2019-tog,choy-2019-cvpr,zhao-2021-iccv,lai-2022-cvpr,nie-2022-cvpr,ye-2022-tpami,zhang-2023-cvpr,li-2023-cvpr,liu-2023-iccv}, a fundamental geometric processing task aiming to assign semantic labels for all points in a scene.
However, even with the most advanced fully-supervised methods, it is hard to achieve consistent high performance on unseen point clouds.
How to handle those mis-segmented regions remains an open issue.
In this work, we focus on interactive point cloud semantic segmentation in an on-the-fly way, which builds on off-the-shelf segmentation networks to enhance their performance with user clicks at the test-time.

Existing interactive segmentation approaches mainly focus on interactive object segmentation (IOS) for images \cite{xu-2016-cvpr,jang-2019-cvpr,kontogianni-2020-eccv,liu-2022-eccv,wei-2023-cvpr} and point clouds \cite{kontogianni-2023-icra,sun-2023-tgrs,yue-2024-iclr} in a customized way, relying on a customized network that receives user clicks and requires offline training.
Interactive single-object segmentation methods \cite{kontogianni-2023-icra,sun-2023-tgrs} receive positive (foreground) and negative (background) clicks as additional features to the input scene, and extracts the object of interest through a customized network, as shown in \cref{fig:intro_ISOS}.
Interactive multi-object segmentation methods \cite{rana-2023-iccv, yue-2024-iclr} treat object clicks as queries to point features, and extract multiple objects simultaneously through a customized network, as illustrated in \cref{fig:intro_IMOS}.
However, both customized approaches above are designed for the IOS task and hard to adapt to the semantic segmentation, since plenty of interactions are required to assign semantic labels for all points in a scene.
Moreover, those customized methods highly rely on specific structures and offline training with tailored supervision, making them decoupled with off-the-shelf segmentation networks and inflexible in practice.

To date, no one has explored the on-the-fly interactive segmentation in the field of point clouds.
Few approaches \cite{wang-2018-tmi,kontogianni-2020-eccv} focus on image IOS in the on-the-fly way.
Concretely, they leverage user interactions as the main supervision, and generate auxiliary supervision to adapt the object segmentation network at the test-time.
To achieve effective pseudo object mask, Wang et al. \cite{wang-2018-tmi} formulate the segmentation refinement process as a conditional random field (CRF) problem, which is solve by Graph Cuts \cite{boykov-2001-iccv}.
However, such a Graph Cut-based serves as a customized algorithm for binary object segmentation and is not suitable for multi-class semantic segmentation in point clouds.
Kontogianni et al. \cite{kontogianni-2020-eccv} directly exploit the initial mask given by a CNN model.
Nevertheless, such auxiliary supervision would introduce a substantial amount of noisy labels, resulting in limited performance, particularly in the more complex point cloud semantic segmentation scenario.

In this paper, we introduce InterPCSeg, the first interactive semantic segmentation (ISS) framework for point clouds in an on-the-fly way.
The InterPCSeg seamlessly integrates with off-the-shelf semantic segmentation networks to enhance their performance at the test-time, as shown in \cref{fig:intro_ISS}.
Specifically, we design a novel test-time loss, consisting of a correction energy and a stabilization energy, to optimize the off-the-shelf networks at the test-time guided by users.
The correction energy, defined on the annotated points, enforces the segmentation refinement based on user corrective clicks, while the stabilization energy, defined on the whole points, stabilizes the refinement process.
To mitigate the inhibitory impact of the stabilization energy on label correction, we further develop a filtering scheme to exclude uncertain points from the stabilization energy.
To objectively and reproducibly evaluate the performance, we propose a novel interaction simulation scheme for the ISS task, which provides corrective clicks like real users.
We conduct experiments on the S3DIS \cite{armeni-2016-cvpr} and ScanNet \cite{dai-2017-cvpr} datasets with two representative backbone networks, working with interactions from real users and the proposed simulator.
Experimental results show that our framework achieves consistent and stable improvement over the baseline with both real and simulated corrective clicks.
The contributions of this work are listed below:
\begin{itemize}
\item We introduce the first interactive semantic segmentation (ISS) framework for point clouds that builds upon off-the-shelf semantic segmentation networks without offline re-training.
\item We establish a novel test-time training scheme by simultaneously minimizing a correction energy and a stabilization energy to facilitate segmentation refinement while maintaining stability.
\item We release an innovative interaction simulator for objective and reproducible performance evaluation for point cloud ISS.
\end{itemize}

\section{Related Works}
\label{sec:related_works}

\subsection{Point cloud semantic segmentation.}
Point cloud semantic segmentation aims to assign per-point semantic labels for point clouds.
The voxel-based \cite{wang-2017-tog,choy-2019-cvpr,zhang-2020-eccv,ye-2022-eccv} and point-based \cite{qi-2017-cvpr,wang-2019-tog,zhao-2021-iccv} methods are the two mainstreams in point cloud semantic segmentation.
The point-based methods directly consume unordered raw point clouds as input and have attracted more attention.
The voxel-based methods convert unordered point clouds into regular voxels, enabling the direct application of 3D convolutional networks.
A comprehensive survey of point cloud semantic segmentation is shown in \cite{guo-2021-tpami}.
Despite the notable advancements of current point cloud semantic segmentation methods, there remains a gap between them and practical applications. 
The open-loop structure of these methods poses limitations when handling distribution shifts in test data.
In this study, we propose an interactive framework to convert these methods into closed-loop systems.

\subsection{Interactive object segmentation.}
Interactive object segmentation (IOS) aims to segment objects of interest guided by user interactions.
Traditional studies solve IOS via heuristic algorithms, such as graph-cut \cite{boykov-2001-iccv,golovinskiy-2009-iccv}, random walk \cite{grady-2006-tpami} and region growing \cite{adams-1994-tpami}.
These studies rely on handcrafted features and demonstrate limited generality when it comes to segmenting visually similar but semantically distinct parts.
Recent works attempt to integrate the interactive scheme with deep neural networks to capture high-level semantics.
Most of them concentrate on IOS for 2D images \cite{xu-2016-cvpr, jang-2019-cvpr, liu-2022-eccv, lin-2022-cvpr-a, du-2023-cvpr, zhou-2023-cvpr}, and few recent works extend IOS to 3D point clouds \cite{kontogianni-2023-icra,sun-2023-tgrs}.

\textbf{Image IOS.}
Based on the action form of user interaction, existing works for image IOS can be divided into the customized way and the on-the-fly way.

Customized methods view interactions as signals to pinpoint the object's location. 
Xu et al. \cite{xu-2016-cvpr} take the lead in the customized IOS by embedding user clicks as additional feature channels to the RGB channels.
To adapt the network to user clicks, they generate many of (image, simulated clicks) input pairs for network re-training.
Recently, numerous studies \cite{mahadevan-2018-bmvc, jang-2019-cvpr, sofiiuk-2020-cvpr, lin-2020-cvpr, zhang-2020-cvpr, hao-2021-iccv, lin-2022-cvpr, chen-2022-cvpr, liu-2022-eccv, lin-2022-cvpr-a, wei-2023-cvpr, huang-2023-iccv, rana-2023-iccv, du-2023-cvpr, zhou-2023-cvpr} follow the pipeline of \cite{xu-2016-cvpr} to improve the effectiveness and efficiency of the customized IOS from different perspectives.
Nevertheless, since the customized IOS methods reform both the input and output structure of the segmentation network, they require offline re-training by densely annotated images.
In addition, the customized IOS methods lacks efficiency in handling semantic segmentation of large scene, as they start segmentation from scratch.

On-the-fly methods \cite{wang-2018-tmi,kontogianni-2020-eccv} take user corrective clicks as supervision for image-specific fine-tuning.
To alleviate the sparsity of supervision, Wang et al. \cite{wang-2018-tmi} utilize the pseudo mask refined by Graph Cuts \cite{boykov-2001-iccv} to guide network adaptation. 
Kontogianni et al. \cite{kontogianni-2020-eccv} leverage the initial segmentation result as the pseudo mask for network adaptation.
Nevertheless, such static pseudo labeling would involve significant poisonous supervision.
To summarize, although the on-the-fly approach provides great flexibility in integrating with existing segmentation networks, there is currently a shortage of suitable methods to address label sparsity. 

\textbf{Point cloud IOS.}
Point cloud IOS received less attention in the past few years compared to its 2D counterpart.
Some studies \cite{kontogianni-2023-icra,sun-2023-tgrs} extend the classic scheme of customized image ISOS to point clouds.
Recently, Yue et al. \cite{yue-2024-iclr} further explore the IMOS task from the perspective of information retrieval.
However, these customized methods demand densely annotated data for retraining and face challenges when extended to the scene-level semantic segmentation task.
In this work, we concentrate on the more challenging and practical task, i.e., point cloud ISS, which aims to assign semantic labels to all points in a scene instead of the objects of interest.

\subsection{Test-time adaptation.}
Test-time adaptation (TTA) aims to adapt a pre-trained network to unlabeled data during the test-time.
A complete review of TTA could be found in \cite{liang-2023-arxiv}.
Among the various categories of TTA techniques, test-time batch adaptation (TTBA) is the most relevant to our framework.
We refer to a special case of TTBA that adapts off-the-shelf networks to a single instance by model optimization.
The key of model optimization is to set appropriate optimization objectives.
Some works \cite{boudiaf-2022-cvpr,zhang-2022-nips} introduce a self-supervised auxiliary task so as to obtain self-supervised objectives during the testing phase.
Some other works \cite{wang-2021-iclr,lee-2023-iccv} directly fine-tune the off-the-shelf networks with unsupervised task-related objectives.
However, both self-supervised and unsupervised objectives fail to provide reliable constraints in the high-precision scenarios.
In this work, we resort to user interactions to establish more reliable and targeted objectives.

\section{Method}
\label{sec:method}

Given a pre-trained point cloud semantic segmentation network $f_{\theta}(.)$ and an unseen point cloud $X \in \mathbb{R}^{N\times C}$, we propose a general framework that leverages user interactions to refine the segmentation result $f_{\theta}(X)$ on-the-fly.

\subsection{Overview}
\begin{figure*}[t]
  \centering
   \includegraphics[width=.9\textwidth]{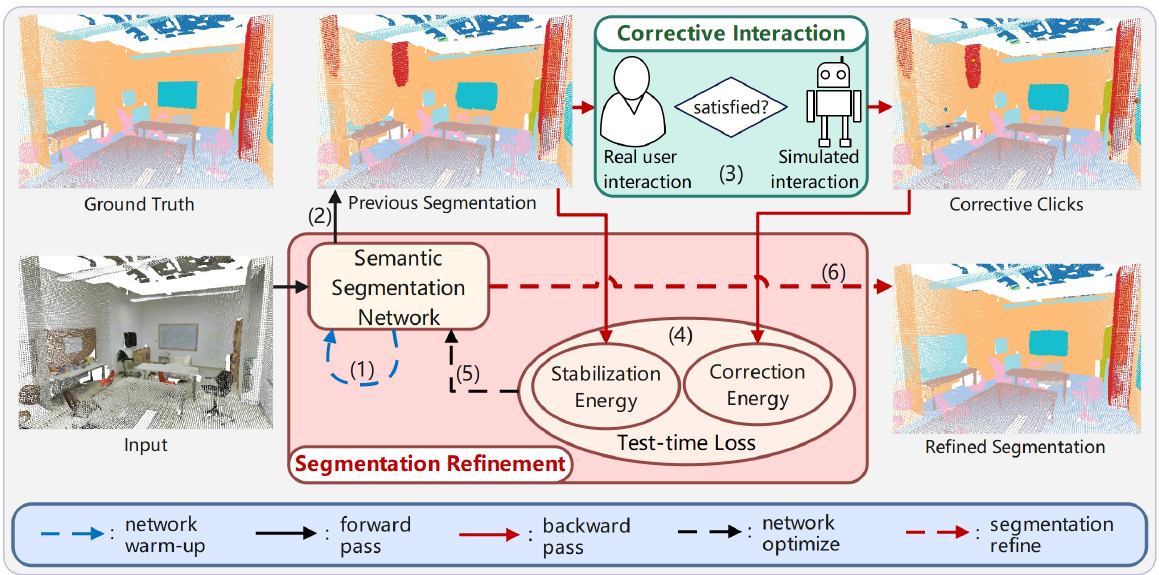}
   \caption{Overview of the InterPCSeg. The pipeline is divided into several steps: (1) Warm up the off-the-shelf semantic segmentation network; (2) Infer with the warmed network; (3) User assesses the current segmentation result and provide corrective clicks or complete the whole process; (4) Calculate the test-time loss based on user clicks and the current segmentation result; (5) Optimize the network parameters; (6) Refine the segmentation result by re-inference.}
   \label{fig:overview}
\end{figure*}

The proposed framework, as shown in \cref{fig:overview}, integrates a pre-trained segmentation network with user interactions on-the-fly to refine the segmentation result through test-time training.
The core components of our framework include a corrective interaction module and a segmentation refinement module. 
These two modules interact with each other to enhance the segmentation result iteratively.

We first introduce a pre-process to warm-up the pre-trained network to alleviate the impact of incorrect batch normalization (BN) statistics during test-time training.
After that, the warmed network infers on the input point cloud, obtaining the current segmentation result.
Then, the user, either real one or simulated one, clicks few points in those obviously mis-segmented parts, and corrects their labels.
Subsequently, the network is trained for few rounds with the test-time constraint, consisting of a correction energy and a stabilization energy.
Finally, once the training is complete, the refined segmentation result would be sent back to the user who determines whether to proceed with further clicks.

\subsection{Network pre-processing}
The network pre-processing is introduced to alleviate the negative impact of incorrect BN statistics in test-time training.

\textbf{Preliminary: BN statistics issue.}
BN layers are widely used in almost all deep networks due to its ability to improve accuracy and speed up training \cite{bjorck-2018-nips}.
A typical BN layer can be expressed as:
\begin{equation}
  y = \gamma  \frac{x-\mu}{\sqrt{\sigma^2 + \epsilon}} + \beta,
  \label{eq:bn}
\end{equation}
where ($x$, $y$) are the input and output of the BN layer, ($\mu$, $\sigma$) are the statistics indicating the mean and standard deviation of $x$, ($\gamma$, $\beta$) are a couple of learnable affine parameters, and $\epsilon$ is set for numerical stability.

In the typical setting, the BN statistics are evaluated and incrementally updated on each mini data batch during training.
During testing, the BN statistics are kept fixed.
While in our setting, during testing, we would like to adapt the pre-trained network to a specific data (i.e., the batch size equals to 1).
To ensure effective training, we relax the BN statistics, allowing them to be evaluated on the specific input data.
However, this operation would introduce biased BN statistics, resulting in degraded initial segmentation results.

\textbf{Network warm-up.}
To ease the BN statistics issue mentioned above, we propose a warm-up process before interaction.
The proposed warm-up process could be divided into the following steps:
First, we conduct inference using the empirical BN statistics incrementally updated on the train set, and recored the segmentation result.
Then, we allow the BN statistics to be evaluated on the specific input data.
Finally, we consider the initial segmentation result as self-supervision to fine-tune the network with the warm-up loss below:
\begin{equation}
  L_{warm-up} = - \frac1N \sum_{i=1}^N \sum_{m=1}^{M} \hat{q}(m|X_i) \mathrm{log}\ p(m|X_i),
  \label{eq:loss_warmup}
\end{equation}
where $\hat{q}(m|X_i) \in \{0,1\}^{N\times M}$ denotes the one-hot encoded pseudo label of $X_i$ given by the original inference result, $p(m|X_i)$ is the probability of $X_i$ being identified as class $m \in \{1,...,M\}$.

\subsection{Corrective interaction} \label{sec:corrective_interaction}
The role of the corrective interaction module is to assess the current segmentation results and provide corrective clicks for refinement.
Typically, the interactions are provided by real users.
Considering the task discrepancy between ISS and IOS, the established interaction simulation methods \cite{kontogianni-2023-icra,sun-2023-tgrs,yue-2024-iclr} designed for the IOS task cannot be generalized to the ISS task.
In order to objectively evaluate our framework and ensure the reproducibility of experiments, we introduce a novel interaction simulation scheme for the ISS task by imitating the behavior of real users.

\textbf{Real user interaction.}
In the task of interactive point cloud semantic segmentation, users should first assess whether the current segmentation result meets their expectations.
Once the current result has met the expectation of user, the segmentation process is complete.
Otherwise, we demand users to prioritize selecting the most obvious mis-segmented regions using relatively inner clicks, and correct the labels of those clicked points.

\begin{figure*}[t]
  \centering
   \includegraphics[width=1.0\textwidth]{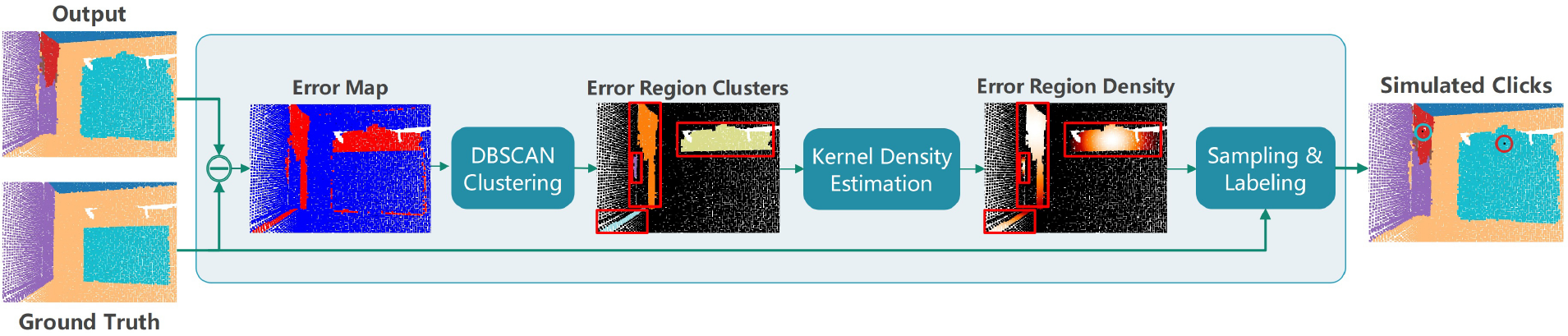}
   \caption{Illustration of our proposed interaction simulation scheme.
   The simulated interactions are achieved in the following steps: (1) Calculate the error map by subtracting the segmentation output from the ground truth; (2) Cluster the error regions to obtain the obvious error regions; (3) Point density estimation on obvious error regions to omit the boundary points; (4) Sample and label the points of interest based on their density.}
   \label{fig:simulated_interaction}
\end{figure*}

\textbf{Simulated interaction.}
We propose a novel interaction simulation scheme as illustrated in \cref{fig:simulated_interaction}, for objective and reproducible experimental analysis of ISS.
The key step is to locate the error regions to interact.
In ISS, the error regions distributed throughout a scene instead of an object.
Therefore, we utilize DBSCAN clustering \cite{ester-1996-kdd} to separate the error regions so as to locate the largest one.
After that, we need to find the ideal clicks which are typically located on the inner part of the region (i.e. inner points).
However, the established centroid-based way may fail when facing the region with a non-convex shape, resulting in clicks on the boundary of the region or outside the region, which may not align with real users.
To this end, we sample the inner points through Kernel Density Estimation \cite{parzen-1962-ams}, considering that the inner points always have higher density than the ones around the boundary.
To sum up, the clustering and density estimation technique jointly build a novel interaction simulation scheme for the ISS task.

\subsection{Segmentation Refinement}
We propose to refine the segmentation result through test-time training.
To achieve effective and stable response to interactions, we design a test-time constraint consisting of a correction energy and a stabilization energy.
The correction energy minimization promotes the correction of mis-segmented regions.
The stabilization energy minimization guarantees the global stability of the segmentation refinement, and facilitates the correction process.
To refine the segmentation result, we optimize the network with the proposed test-time constraint.

\textbf{Correction energy.}
The correction energy reflects the fidelity of clicked points being correctly segmented, defined by the sparse cross entropy on those user clicked points:
\begin{equation}
  E_{correction} = - \sum_{i=1}^U \sum_{m=1}^{M} q(m|X_i) \mathrm{log}\ p(m|X_i),
  \label{eq:energy_correction}
\end{equation}
where $U$ represents the number of user clicks, $p(m|X_i)$ is the probability of $X_i$ being identified as class $m \in \{1,...,M\}$, and $q(m|X_i) \in \{0,1\}^{N\times M}$ denotes the one-hot encoded label of $X_i$ given by the user. 
Since all points share the same network weights, minimizing the correction energy on a sparse point set could lead to alterations in the predictions across all points.

\textbf{Stabilization energy.}
The stabilization energy assesses the stability of predictions, which is defined as follows: 
\begin{equation}
  E_{stabilization} = - \frac1N \sum_{i=1}^N \sum_{m=1}^{M} s_i p(m|X_i) \mathrm{log}\ p(m|X_i),
  \label{eq:energy_stabilization}
\end{equation}
where $N$ denotes the number of points, and $s_i$ represents the filtering score of the prediction of $X_i$.
We design a dynamic update rule for the point-wise filtering score $s_i$.
Concretely, we reset the filtering scores after each round of interactions, according to the algorithm presented in \cref{algorithm}.
Furthermore, we update the filtering score in each round of optimization except for the interaction round, according to the entropy change between adjacent optimization rounds:
\begin{equation}
\begin{split}
  &s_i^{t+1} =
  \begin{cases}
  1  & \text{if $\Delta E_i^t  < - \delta^{-}$} \\
  0 & \text{if $\Delta E_i^t  > \delta^{+}$} \\
  s_i^t & \text{else}
  \end{cases}, \\
  & \Delta E_i^t = E_i^{t+1} - E_i^t, \  t=0,...,T,
  \label{eq:certainty_assess}
\end{split}
\end{equation}
where $\Delta E_i^t$ denotes the entropy change between round $t$ and $t+1$,
$\delta^{+}$ and $\delta^{-}$ are pre-defined thresholds, 
and $T$ is the number of optimization rounds after each interaction.

\begin{algorithm}[!t]
\label{algorithm}
\DontPrintSemicolon
\caption{Filtering Score Evaluation}\label{algorithm}
\KwData{input $X \in \mathbb{R}^{N\times C}$, segmentation network $f_{\theta}$, hyper-parameters $\lambda, \delta$}
\KwResult{filtering score $s_i, i=1...N$}
$s_i \leftarrow 1, \text{for}\ i=1,...N$\  \tcp*{initialize $s_i$}
$\hat{f}_{\theta} \leftarrow f_{\theta}.\text{clone()}$\	\tcp*{network copy}
$p \leftarrow \hat{f}_{\theta}(X)$\;
$E_i \leftarrow \sum_{m=1}^{M}p(m|X_i) \mathrm{log}\ p(m|X_i), \text{for}\ i=1,...,N$\ \tcp*{record entropy}
$\text{compute}\ E_{correction}, E_{stabilization}$
$\text{based on \cref{eq:energy_correction} and \cref{eq:energy_stabilization}}$\;
$\text{loss} \leftarrow E_{correction}+\lambda E_{stabilization}$\;
$\text{loss.backward()}$\;
$\text{update}(\hat{f}_{\theta}\text{.params})$\	\tcp*{update network parameters}
$p^{\prime} \leftarrow \hat{f}_{\theta}(X)$\;
$E^{\prime}_i \leftarrow \sum_{m=1}^{M}p^{\prime}(m|X_i) \mathrm{log}\ p^{\prime}(m|X_i),$
$\text{for}\ i=1,...,N$\;
$\Delta E_i \leftarrow E_i^{\prime} - E_i,\text{for}\ i=1,...,N$\ \tcp*{calculate change of entropy}
$s_i \leftarrow 0 \ \  \text{if}\ \  \Delta E_i \geq \delta\ \ \text{else} \ \ 1,\text{for}\ i=1,...,N$\ \tcp*{update $s_i$}
\end{algorithm}

\textbf{Network optimization.}
We formulate the test-time loss below to enforce network optimization:
\begin{equation}
  L_{test-time} = E_{correction} + \lambda E_{stabilization},
  \label{eq:loss_test-time}
\end{equation}
where $\lambda$ is a trade-off factor.
We utilize the same optimizer as used in training of the backbone networks, but we omit the gradient accumulation (GA) process in the test-time training.
The GA is typically used as an approach to accelerate the convergence in several optimizers, such as the stochastic gradient descent (SGD) optimizer \cite{kiefer-1952-ams} and the adaptive moment estimation (Adam) optimizer \cite{kingma-2014-arxiv}.
However, in our setting, the optimization objective is dynamic, so it would introduce instability in the test-time training stage.
The test-time training would be continued for few rounds to ensure that the segmentation result is completely stabilized.
After that, the segmentation result would be sent back to the user.

\section{Experiment}
\label{sec:experiment}

\subsection{Experiment Configuration}
\textbf{Datasets.}
We evaluate our framework on two popular indoor point cloud datasets, including S3DIS \cite{armeni-2016-cvpr} and ScanNet \cite{dai-2017-cvpr}.
The S3DIS comprises 6 areas consisting of 272 scenes with 13 categories.
We pick scenes in Area 5 for interactive segmentation, while the remaining scenes are used for training the backbone network.
Considering the computational cost, each scene is subsampled with a 3 cm grid.
For few large scenes that contain numerous points ($\ge$ 150k points) after subsampling, we further crop them into sub-scenes along the longest axis for better handling.
The ScanNet contains 1513 scenes and covers 20 semantic categories.
We follow the strategy in \cite{dai-2017-cvpr} for train-test splitting.
Note that our proposed interactive framework is only conducted on the test set, equipped with a backbone segmentation network pre-trained on the train set.

\textbf{Backbone networks.}
We directly utilize the off-the-shelf networks and pre-trained parameters provided by the Open3D library \cite{zhou-2018-arXiv} without making any modification.
Specifically, we employ the PointTransformer (PT) \cite{zhao-2021-iccv} which is pre-trained on the S3DIS dataset as well as the SparseConvUnet (SCU) \cite{choy-2019-cvpr} which is pre-trained on the ScanNet dataset.
Both of them are representative works in point-based and voxel-based point cloud semantic segmentation, respectively.
Note that other off-the-shelf networks can also be used as the backbone networks.

\textbf{Evaluation metrics.}
We employ the mean Intersection over Union (mIoU) as the metric to evaluate the segmentation result.
To assess the effectiveness of InterPCSeg, we measure two key metrics: the Number of Clicks (NoC) required to achieve a predefined mIoU value, and the mIoU obtained with a predetermined NoC.
Considering that our method builds upon a baseline off-the-shelf network, we mainly focus on the improvement over the baseline instead of the absolute performance.

\textbf{Interactions.}
For fair evaluation in quantitative analysis, we leverage the proposed simulated interaction algorithm (refer to \cref{sec:corrective_interaction}) to generate corrective clicks.
While in qualitative analysis, we recruit volunteers for interaction to show the effectiveness in real-world scenarios.
Furthermore, we compare the simulated interaction with that from real users quantitatively and qualitatively respectively.
The interaction budget is set to 30 clicks in our experiments.
If the current segmentation result has met the requirements before reaching the predefined interaction budget, the interaction process would be terminated early.

\textbf{Comparative methods.}
Since the InterPCSeg is the first attempt to leverage user interactions to enhance the performance of off-the-shelf semantic segmentation networks, there are no existing works available for direct comparison.
Note that the customized IOS methods are decoupled with off-the-shelf networks and require offline supervised training.
Furthermore, it is time-consuming to apply those customized IOS methods to semantic segmentation, since they start segmentation from scratch and require interacting with objects one by one to achieve semantic segmentation.
Therefore, we refer to the on-the-fly IOS method.
BIFSeg \cite{wang-2018-tmi} is hard to apply to the semantic segmentation scenario, since it depends on pseudo labeling by Graph Cuts \cite{boykov-2001-iccv}, which only works for binary object segmentation and is time-consuming for large-scale point clouds.
IA \cite{kontogianni-2020-eccv} directly works with off-the-shelf networks without any task-oriented process, and thus can be extended to our setting.

\textbf{Implementation details.}
For the PT, we set $\delta = \delta^{+} = \delta^{-} = 0.03$ for confidence score evaluation, $\lambda=100$ for test-time loss calculation. The learning rate is set to $5e-3$ for network warm-up and $1e-3$ in other scenarios.
We utilize the standard SGD optimizer with momentum of 0.9 and weight decay of 0.01 in network warm-up and confidence score evaluation.
During the test-time training, the momentum is removed.
We perform optimization of 5 rounds for the warm-up stage and 3 rounds for the test-time training after receiving interactions.

For the SCU, we set $\delta = \delta^{+} = 0.1, \delta^{-}=0.01$, $\lambda=100$.
We employ the Adam optimizer with $\beta_1=0.9, \beta_2=0.999$ and weight decay of 0.5 in network warm-up and confidence score evaluation. The $\beta_1$ and $\beta_2$ are set to zero in test-time training.
We take 10 rounds of optimization for the network warm-up and 5 rounds for the test-time training.


\begin{table}[t]
  \centering
  \caption{NoC needed to achieve the target mIoU as well as the failure rate within the limited interaction budget (30 clicks).}
  \setlength{\tabcolsep}{1.5mm}{
  \begin{tabular*}{1.0\textwidth}{@{\extracolsep{\fill}} l | c c | c c | c c | c c}
    \toprule
    \multirow{3}{*}{Methods} &
    \multicolumn{4}{c|}{PT@S3DIS (base mIoU=65\%)} &
    \multicolumn{4}{c}{SCU@ScanNet (base mIoU=55\%)} \\
     &	\multicolumn{2}{c|}{85\%mIoU} & 
	\multicolumn{2}{c|}{90\%mIoU} &
	\multicolumn{2}{c|}{80\%mIoU} & 
	\multicolumn{2}{c}{85\%mIoU} \\
     & NoC & Failures & NoC & Failures &
     NoC & Failures & NoC & Failures \\
    \midrule
    IA \cite{kontogianni-2020-eccv} & 25.9 & 80.0\% & 28.1 & 85.3\% & 24.4 & 75.3\% & 26.9 & 86.4\% \\
    w/o GA removal & 18.2 & 56.0\% & 24.2 & 74.7\% & 11.8 & 17.5\% & 15.3 & 31.8\% \\
    w/o stabilization & 20.9 & 46.7\% & 24.0 & 54.7\% & 24.8 & 76.9\% & 26.8 & 89.3\% \\
    w/o filtering & 10.2 & 16.0\% & 13.8 & 21.3\%  & 12.4 & 18.5\% & 16.3 & 33.4\%\\
    w/o warm-up & 10.9 & \textbf{13.3\%} & 15.2 & \textbf{17.3\%} & 11.4& 15.6\% & 15.4 & 29.5\% \\
    \textbf{Full pipeline} & \textbf{9.5} & 16.0\% & \textbf{13.3} & 20.0\% & \textbf{11.1} & \textbf{14.3\%} & \textbf{14.2} & \textbf{25.3\%} \\
    \bottomrule
  \end{tabular*}}
  \label{tab:avg_perform}
\end{table}

\subsection{Results} \label{exp:results}
\textbf{Quantitative analysis.}
\cref{tab:avg_perform} reports the average NoC needed to achieve the target mIoU and records the number of failure cases within the interaction budget.
Given the baseline networks achieving 65\% mIoU on the S3DIS and 55\% mIoU on the ScanNet respectively, our full pipeline increases the performance remarkably with a few user clicks.
Specifically, for S3DIS, our full framework achieves 85\% mIoU using 9.5 clicks by average for 84\% scenes, and 90\% mIoU using 13.3 clicks by average for 80\% scenes.
For ScanNet, our full framework obtains 80\% mIoU using 11.1 clicks by average for 85.7\% scenes, and 85\% mIoU using 14.2 clicks by average for 74.7\% scenes.
When it comes to IA \cite{kontogianni-2020-eccv}, 80\% and 85.3\% scenes of S3DIS fail to be refined to the mIoUs of 85\% and 90\%, respectively.
For ScanNet, the failure rates of IA are 75.3\% for the mIoU of 80\%, and 86.4\% for the mIoU of 85\%, respectively.

\begin{figure}[t]
  \centering
  \begin{subfigure}{0.3\linewidth}
    \includegraphics[width=1\textwidth]{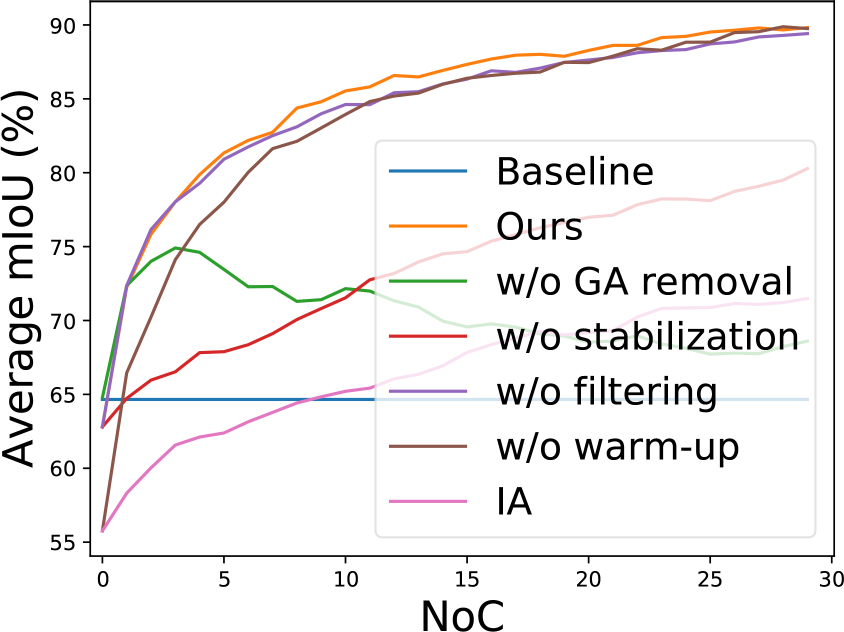}
    \caption{Average performance on PT@S3DIS\\}
    \label{fig:avg_perform_s3dis}
  \end{subfigure}
  \begin{subfigure}{0.3\linewidth}
    \includegraphics[width=1\textwidth]{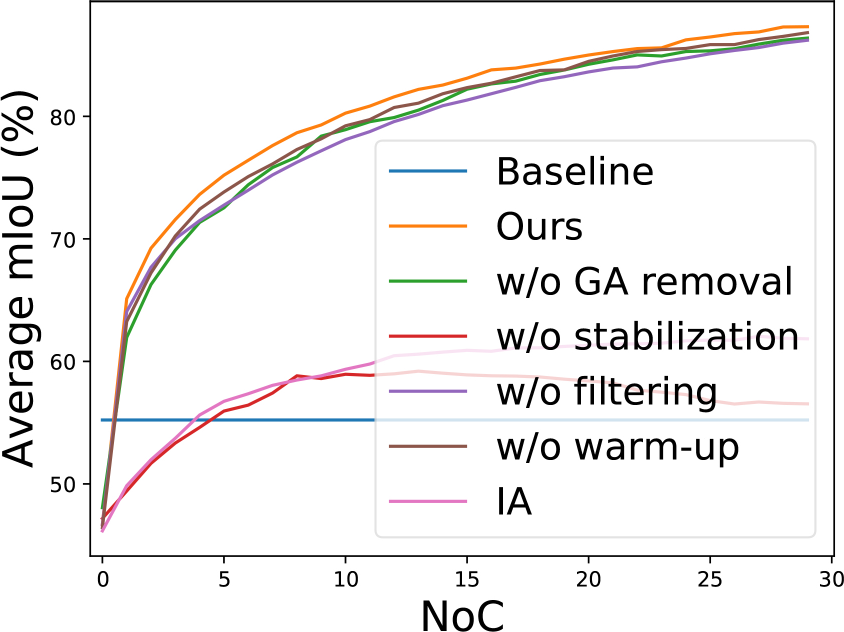}
    \caption{Average performance on SCU@ScanNet}
    \label{fig:avg_perform_scannet}
  \end{subfigure}
   \begin{subfigure}{0.3\linewidth}
    \includegraphics[width=1\textwidth]{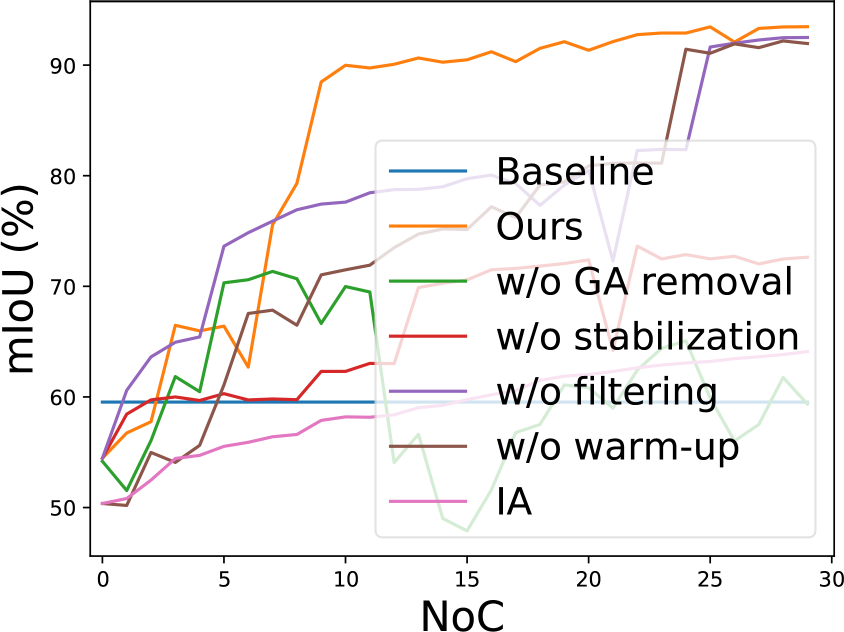}
    \caption{Instance performance on PT@S3DIS}
    \label{fig:hard_perform_s3dis}
  \end{subfigure}
  \begin{subfigure}{0.3\linewidth}
    \includegraphics[width=1\textwidth]{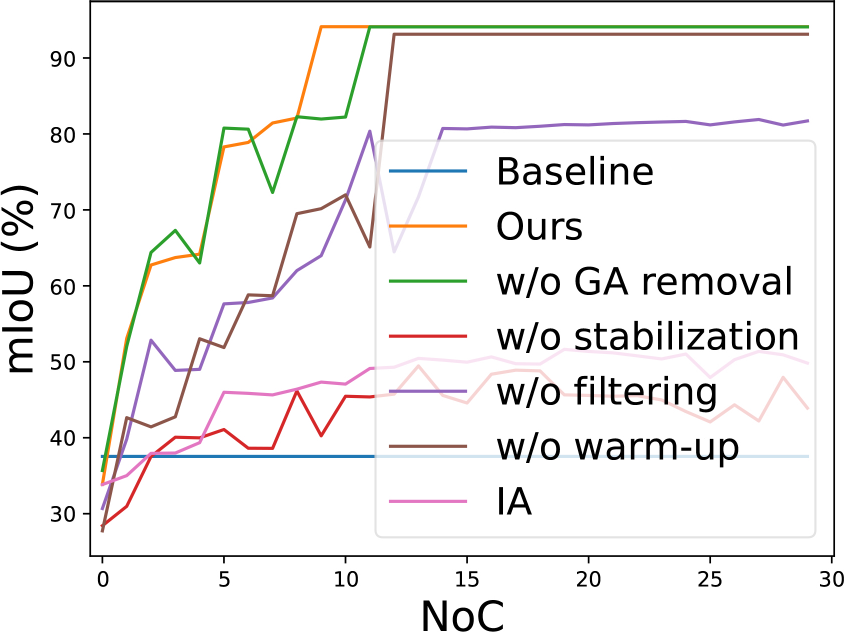}
    \caption{Instance performance on SCU@ScanNet}
    \label{fig:hard_perform_scannet}
  \end{subfigure}
  \begin{subfigure}{0.3\linewidth}
    \includegraphics[width=1\textwidth]{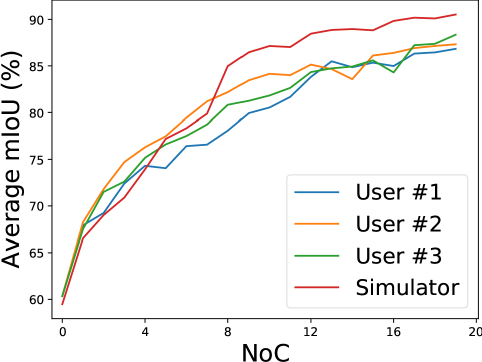}
    \caption{User interaction studies on PT@S3DIS}
    \label{fig:real_user_s3dis}
  \end{subfigure}
  \begin{subfigure}{0.3\linewidth}
    \includegraphics[width=1\textwidth]{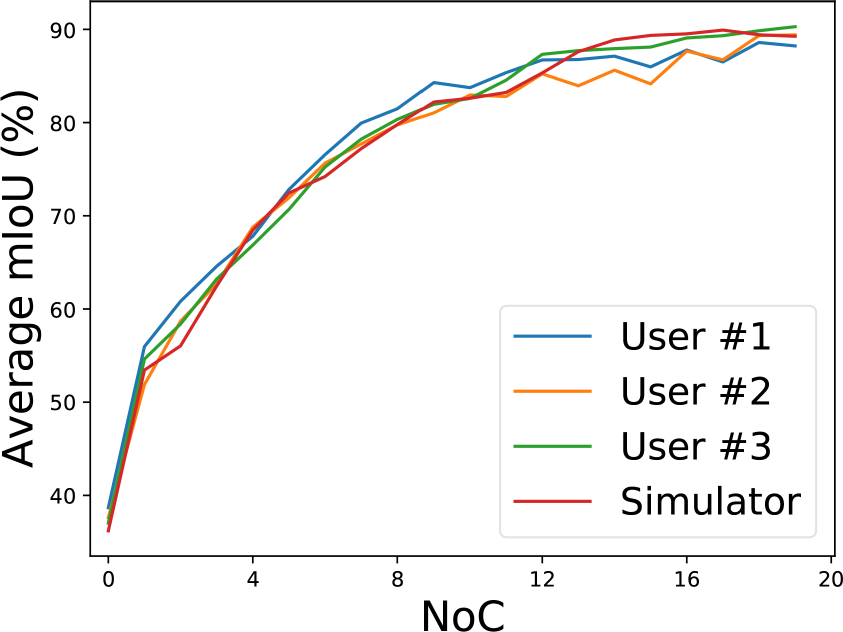}
    \caption{User interaction studies on SCU@ScanNet}
    \label{fig:real_user_scannet}
  \end{subfigure}
  \caption{The mIoU curves with respect to the number of clicks (NoC).}
  \label{fig:exp_quanti}
\end{figure}

\cref{fig:avg_perform_s3dis,fig:avg_perform_scannet} show the average mIoU curves with respect to NoC.
Our full framework demonstrates a similar logarithmic-like increase in mIoU as the NoC increases.
Ultimately, the full pipeline achieves more than 25 \% mIoU improvement over the baseline within 30 clicks for both the S3DIS and the ScanNet.
In contrast, the IA \cite{kontogianni-2020-eccv} achieves less than 10\% mIoU improvement within 30 clicks across the two datasets.
To better reveal the evolutionary process of interactive segmentation, we extract two challenging instances for quantitative evaluation.
As shown in \cref{fig:hard_perform_s3dis,fig:hard_perform_scannet}, the full pipeline obtains substantial performance with a minimal NoC and stable improvement with additional clicks.

\textbf{Real users v.s. interaction simulator.}
To evaluate our framework under real user interactions, we recruit three volunteers and ask them to annotate on 10 scenes from the S3DIS and 15 scenes from the ScanNet.
\cref{fig:real_user_s3dis,fig:real_user_scannet} report the average mIoU values within 20 real user clicks.
We additionally include the interaction simulator for comparison.

In \cref{fig:real_user_s3dis}, the interaction simulator achieves even better performance than real users.
This phenomenon can be attributed to the fact that real users typically focus on the local region where they are interacting, while ignoring other potentially valuable regions outside of their immediate view.
While in the ScanNet where the scene scale is relatively small, this phenomenon would disappear, as shown in \cref{fig:real_user_scannet}.
In total, our proposed interaction simulation scheme achieves similar performance to real users.

\begin{figure*}[!t]
  \centering
  \includegraphics[width=0.9\textwidth]{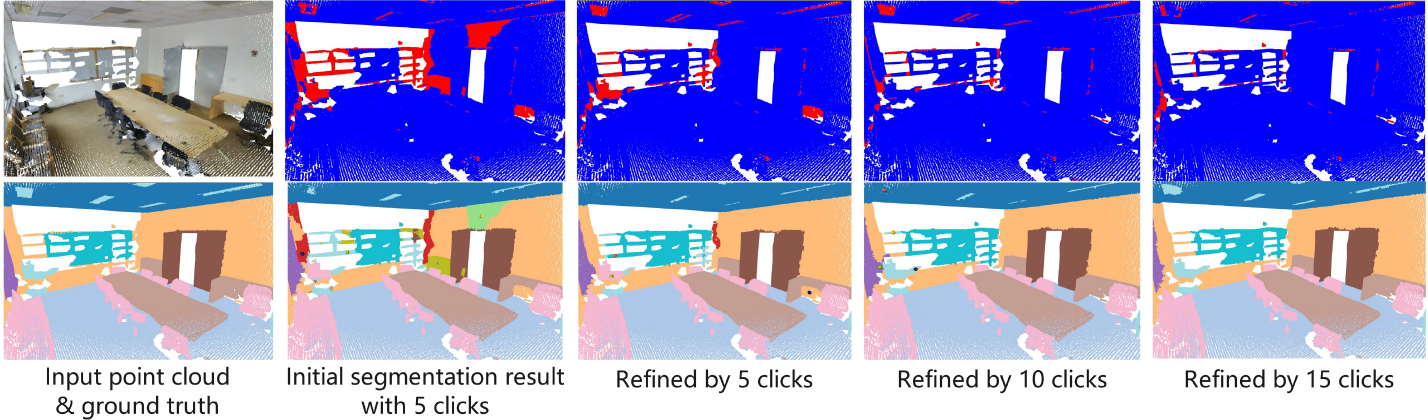}
  \caption{An interactive segmentation process by our proposed framework.
  The top row represents the input point cloud and the error map of each segmentation result. 
  The bottom row consists of the ground truth result, initial segmentation result, and the refined results.
  The corrective clicks (totally 15 clicks) are progressively provided, marked as colored dots on the initial and refined segmentation results.}
  \label{fig:exp_quality}
\end{figure*}

\textbf{Qualitative analysis.}
\cref{fig:exp_quality} presents a complete process of interactive segmentation refinement with our framework.
The primary error regions are removed by 5 clicks on the initial segmentation result.
After that, 10 extra clicks are provided to correct secondary errors and perform local refinement.
The initial segmentation result could be globally refined within few rounds of clicks.

\begin{figure*}[t]
  \centering
  \includegraphics[width=0.9\textwidth]{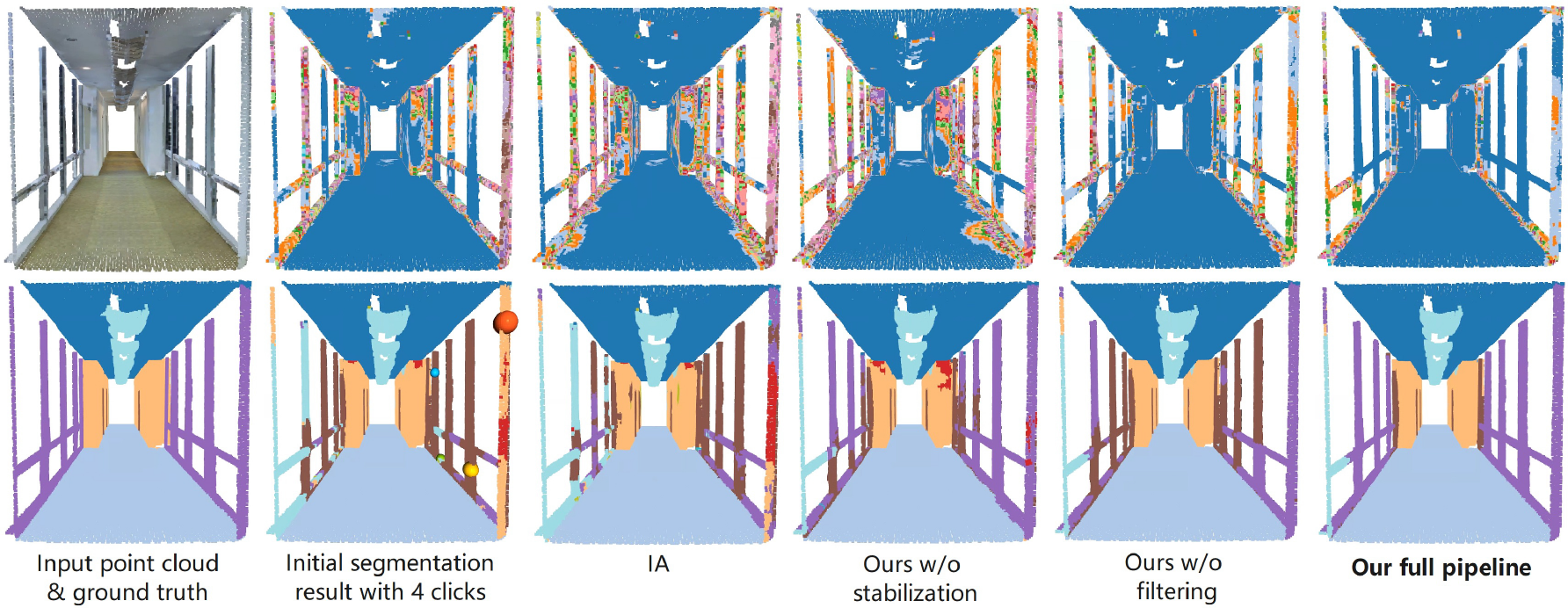}
  \caption{Qualitative analysis on a special case of PT@S3DIS.
  The top row shows the input point cloud and the entropy heat map of each segmentation result.
  The bottom row contains the ground truth result, the initial segmentation result, and refined results.}
  \label{fig:exp_quality_ablation}
\end{figure*}

In \cref{fig:exp_quality_ablation}, we provide a visualization case of segmentation refinement on a challenging point cloud.
The selected point cloud owns a symmetrical structure and the corrective clicks concentrate on one side of the symmetrical scene.
We provide 4 clicks in one pass for segmentation refinement.
It is observable that our full framework refines the mis-segmented parts not only on the side we click, but also on the other side where no clicks are provided.
Furthermore, the segmentation refinement remains stable, as shown in the entropy map in \cref{fig:exp_quality_ablation} (top row).
On the contrary, the IA \cite{kontogianni-2020-eccv} fails to refine the clicked mis-segmented regions, and even introduces extra mis-segmentation, demonstrating significant instability as depicted in its entropy map in \cref{fig:exp_quality_ablation} (top row).

\textbf{Running time analysis.}
The running time of our framework is determined by various factors, including the complexity of the backbone networks, the number of rounds allocated for test-time training, as well as the scale of the input.
In our setting, it takes 3.54s for SCU to refine a scene of 23760 points, and 5.42s for PT to refine a scene of 68980 points.
Such performance is acceptable, since one can offer several clicks in one round and only a few rounds are required to achieve satisfactory segmentation result in most cases.

\subsection{Ablation Studies} \label{exp:ablation}
We further conduct ablation studies to evaluate the effectiveness of the essential designs in our framework, including stabilization energy, filtering scheme, GA removal and network warm-up. 
The experimental settings remain identical to those for the full pipeline evaluation.
\cref{tab:avg_perform,fig:avg_perform_s3dis,fig:avg_perform_scannet,fig:hard_perform_s3dis,fig:hard_perform_scannet} report the quantitative analysis of the ablation variants.
\cref{fig:exp_quality_ablation} presents a qualitative ablation study of two key designs, i.e., the stabilization energy and filter scheme, on a challenging point cloud.

\textbf{Stabilization energy.}
The stabilization energy ensures stable segmentation refinement.
As shown in \cref{fig:exp_quality_ablation}, when the stabilization energy is removed, the entropy values of parts surrounding clicks increase significantly, resulting in unstable and unreasonable segmentation refinement.
As presented in the those quantitative results, relying solely on the correction energy for test-time training would result in significant performance degradation.

\textbf{Filtering scheme.}
Filtering scheme serves as an embedded component in the stabilization energy to suppress the conflict between the correction and stabilization energy.
As shown in \cref{fig:exp_quality_ablation}, without the filtering scheme, our framework shows under refinement due to the negative impact of stabilization energy, and achieves higher entropy.
The quantitative results also illustrate slight performance decline when the filtering scheme is removed.

\textbf{GA removal.}
GA is removed for test-time training, since the optimization objective in our setting is dynamic, resulting in unstable optimization.
As shown in \cref{fig:avg_perform_s3dis,fig:avg_perform_scannet}, when the GA strategy is recovered, the performance drops especially for PT@S3DIS that is optimized by the SGD optimizer.
The SGD optimizer inherits gradient of the past step, resulting in unstable gradient due to the dynamic optimization objective.
In contrast, the Adam optimizer is able to adaptively adjust the learning rate according to the varying gradient, and thus demonstrates more robust performance to the dynamic optimization objective.

\textbf{Network warm-up.}
Network warm-up serves as a pre-process step to alleviate the impact of incorrect BN statistics during test-time training.
As illustrated in \cref{fig:avg_perform_s3dis}, towards passing the warm-up process, the initial mIoU obviously declines, resulting in an increase of NoC.
Notably, when adequate clicks are provided, the mIoU still reach the same level as the full pipeline, showing the effectiveness of test-time training.

\section{Conclusion}
In this paper, we introduced the first interactive framework for point cloud semantic segmentation, which directly works with the off-the-shelf networks in an on-the-fly way.
In our framework, we treated user corrective clicks as sparse training examples to optimize the network parameters.
To handle instability due to the sparse supervision, we devised a novel stabilization energy that ensures stable segmentation refinement.
Moreover, we developed an interaction simulation scheme to objectively and reproducibly assess the proposed ISS framework.
Extensive experiments show the efficacy of our framework across different backbone networks, datasets, and user scenarios.

\bibliographystyle{splncs04}
\bibliography{references}
\end{document}